\documentclass[11pt,a4paper]{article}
\usepackage[hyperref]{acl2019}
\usepackage{times}
\usepackage{helvet}
\usepackage{courier}
\usepackage{url}
\usepackage{graphicx}
\usepackage{latexsym}
\usepackage{mathptmx}
\usepackage[hyphenbreaks]{breakurl}
\usepackage{xcolor}
\usepackage{bibentry}
\usepackage{stfloats}
\usepackage{subfig}
\usepackage{multirow}
\usepackage{amssymb}
\usepackage{pifont}
\newcommand{\xmark}{\ding{55}}%
\usepackage{amsmath}
\allowdisplaybreaks
\usepackage{array}

\DeclareMathAlphabet{\mathcal}{OMS}{cmsy}{m}{n}
\definecolor{mypink}{rgb}{1., 0, 0}

\aclfinalcopy 

\title{DOER: Dual Cross-Shared RNN for Aspect Term-Polarity Co-Extraction}

\author{Huaishao Luo$^{1}$, Tianrui Li$^{1}$\thanks{\;\;Tianrui Li is the corresponding author.}\,\ , Bing Liu$^2$, Junbo Zhang$^{3,4,5}$\\
  $^1$School of Information Science and Technology, Southwest Jiaotong University, China \\
  {\tt huaishaoluo@gmail.com, trli@swjtu.edu.cn} \\
  $^2$Department of Computer Science, University of Illinois at Chicago, USA \\
  {\tt liub@uic.edu} \\
  $^3$JD Intelligent Cities Business Unit \& $^4$JD Intelligent Cities Research, China \\
  $^5$Institute of Artificial Intelligence, Southwest Jiaotong University, China \\
  {\tt msjunbozhang@outlook.com}
  }

\date{}
\begin{document}
	\maketitle
	\begin{abstract}
		This paper focuses on two related subtasks of aspect-based sentiment analysis, namely aspect term extraction and aspect sentiment classification, which we call \textit{aspect term-polarity co-extraction}. The former task is to extract aspects of a product or service from an opinion document, and the latter is to identify the polarity expressed in the document about these extracted aspects. Most existing algorithms address them as two separate tasks and solve them one by one, or only perform one task, which can be complicated for real applications. In this paper, we treat these two tasks as two sequence labeling problems and propose a novel Dual crOss-sharEd RNN framework (DOER) to generate all aspect term-polarity pairs of the input sentence simultaneously. Specifically, DOER involves a dual recurrent neural network to extract the respective representation of each task, and a cross-shared unit to consider the relationship between them. Experimental results demonstrate that the proposed framework outperforms state-of-the-art baselines on three benchmark datasets.
	\end{abstract}
	
	\section{Introduction}
	\label{sec_introduction}
	Aspect terms extraction (ATE) and aspect sentiment classification (ASC) are two fundamental, fine-grained subtasks of aspect-based sentiment analysis. Aspect term extraction is the task of extracting the attributes (or aspects) of an entity upon which opinions have been expressed, and aspect sentiment classification is the task of identifying the polarities expressed on these extracted aspects in the opinion text \cite{Hu2004}. Consider the example in Figure \ref{fig_examples}, which contains comments that people expressed about the aspect terms ``operating system", ``preloaded software", ``keyboard", ``bag", ``price", and ``service" labeled with their polarities, respectively. The polarities contain four classes, e.g., positive (PO), conflict (CF), neutral (NT)\footnote{Neutral means no sentiment is expressed, and we also regard it as a polarity as in many prior works.}, and negative (NG). 
	
	To facilitate practical applications, our goal is to solve ATE and ASC simultaneously. For easy description and discussion, these two subtasks are referred to as aspect term-polarity co-extraction. Both ATE and ASC have attracted a great of attention among researchers, but they are rarely solved together at the same time due to some challenges:
	
	\begin{figure}
	\centering
		\includegraphics[width=0.48\textwidth, keepaspectratio]{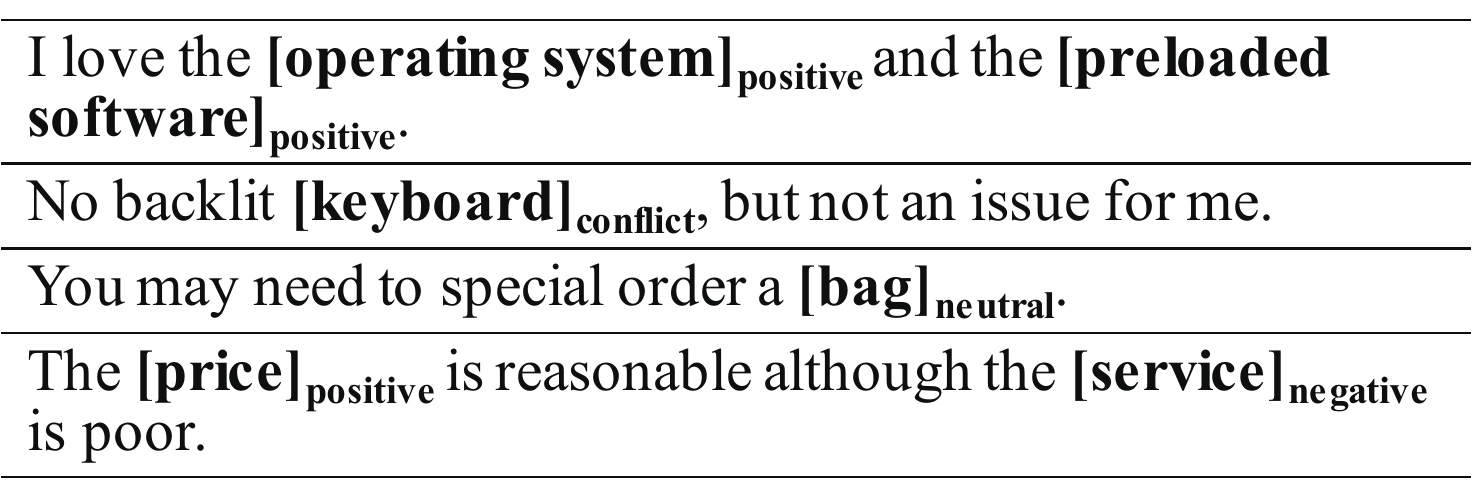}
		\caption{Aspect terms extraction and aspect sentiment classification.}\label{fig_examples}
	\end{figure}
	
	1) \textit{ATE and ASC are quite different tasks.} ATE is an extraction or sequence labeling task~\cite{Jakob2010,Wang2016a}, while ASC is a classification task~\cite{Jiang2011target,Wagner2014,Tang2016a,Tang2016,Tay2018a}. Thus, they are naturally treated as two separate tasks, and solved one by one in a pipeline manner. However, this two-stage framework is complicated and difficult to use in applications because it needs to train two models separately. There is also the latent error propagation when an aspect term is used to classify its corresponding polarity. Thus, due to the different natures of the two tasks, most current works focus either on extracting aspect terms \cite{Yin2016,Luo2018,Xu2018} or on classifying aspect sentiment \cite{Ma2017a,Wang2018}. A possible idea to bridge the difference between the two tasks is to change ASC to a sequence labeling task. Then, ATE and ASC have the same formulation.
	
	2) \textit{The number of aspect term-polarity pairs in a sentence is arbitrary.} Considering the examples depicted in Figure \ref{fig_examples}, we can observe that some sentences contain two term-polarity pairs and some sentences contain one pair. Moreover, each aspect term can consist of any number of words, which makes the co-extraction task difficult to solve. 
	
\begin{figure}[tbp]
		\centering
		\includegraphics[width=0.49\textwidth, keepaspectratio]{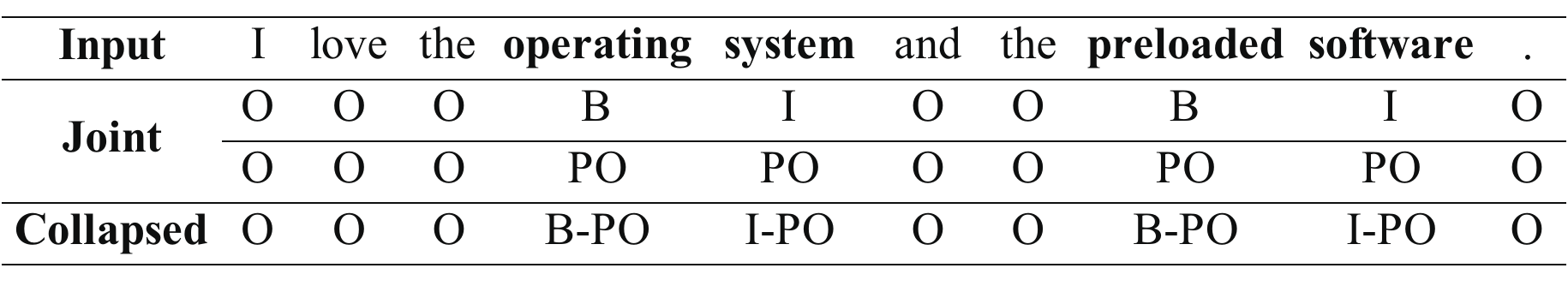}
		\caption{A labeling example of aspect terms and their polarities.}\label{table_labeling_examples}
	\end{figure}

	Some existing research has treated ATE and ASC as two sequence labeling tasks and dealt with them together.
	\citeauthor{Mitchell2013} (\citeyear{Mitchell2013}) and \citeauthor{Zhang2015} (\citeyear{Zhang2015}) compared pipelined, joint, and collapsed approaches to extracting named entities and their sentiments. They found that the joint and collapsed approaches are superior to the pipelined approach. \citeauthor{Li2017}  (\citeyear{Li2017}) proposed a collapsed CRF model. The difference with the standard CRF is that they expanded the node type at each word to capture sentiment scopes. Another interesting work comes from \citeauthor{Li2019} (\citeyear{Li2019}), where the authors proposed a unified model with the collapsed approach to do aspect term-polarity co-extraction. We can intuitively explain the pipelined, joint, and collapsed approaches through Figure \ref{table_labeling_examples}. The pipelined approach first labels the given sentence using aspect term tags, e.g., ``B'' and ``I'' (the Beginning and Inside of an aspect term) and then feeds the aspect terms into a classifier to obtain their corresponding polarities. The collapsed approach uses collapsed labels as the tags set, e.g., ``B-PO'' and ``I-PO''. Each tag indicates the aspect term boundary and its polarity. The joint approach jointly labels each sentence with two different tag sets: aspect term tags and polarity tags.
	
	We believe that the joint approach is more feasible than the collapsed approach when integrating with neural networks because the combined tags of the latter may easily make the learned representation confused. As an example in Figure \ref{table_labeling_examples}, the ``operating system'' is an aspect term. Its polarity ``positive'' actually comes from the word ``love''. They should be learned separately because the meanings of these two groups of words are different. That means that using ``B-PO I-PO'' to extract the meaning of ``operating system'' and ``love'' simultaneously is difficult in training (this will be clearer later). In contrast, the joint approach has separate representations for ATE and ASC and separate labels. Thus, an extra sentiment lexicon can improve the representation of ASC individually, and the interaction of ATE and ASC can further enhance the performance of each other.
	
	In this paper, we propose a novel Dual crOss-sharEd RNN framework (DOER) to generate all aspect term-polarity pairs of a given sentence. DOER mainly contains a dual recurrent neural network (RNN) and a cross-shared unit (CSU). The CSU is designed to take advantage of the interactions between ATE and ASC. Apart from them, two auxiliary tasks, \textit{aspect length enhancement} and \textit{sentiment enhancement}, are integrated to improve the representation of ATE and ASC. An extra RNN cell called the \textit{Residual Gated Unit} (ReGU) is also proposed to improve the performance of aspect term-polarity co-extraction. The ReGU utilizes a gate to transfer the input to the output like skip connection \cite{He2016}, and thus, is capable of training deeper and obtaining more useful features. In a word, DOER generates aspect terms and their polarities simultaneously by an end-to-end method instead of building two separate models, which saves time and gives a unified solution to practical applications.
	
	Our contributions are summarized as follows:
	
	\begin{itemize}
		\item A novel framework DOER is proposed to address the aspect term-polarity co-extraction problem in an end-to-end fashion. A cross-shared unit (CSU) is designed to leverage the interaction of the two tasks.
		\item Two auxiliary tasks are designed to enhance the labeling of ATE and ASC, and an extra RNN cell ReGU is proposed to improve the capability of feature extraction.
	\end{itemize}
	\section{Methodology}
	\label{sec_methodology}
\begin{figure*}[htbp]
		\centering
		\subfloat[Dual cross-shared RNN framework (DOER)] {
			\centering
			\includegraphics[width=0.68\textwidth, keepaspectratio]{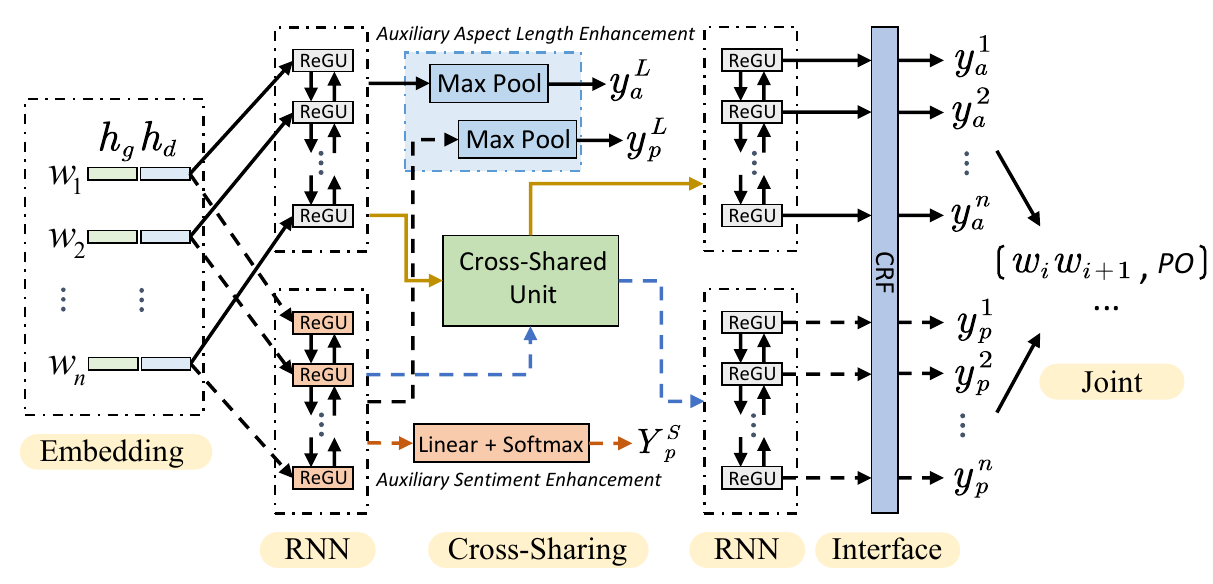}
			\label{fig_framework_dcs}
		}
		\subfloat[Cross-shared unit (CSU)] {
			\centering
			\includegraphics[width=0.30\textwidth, keepaspectratio]{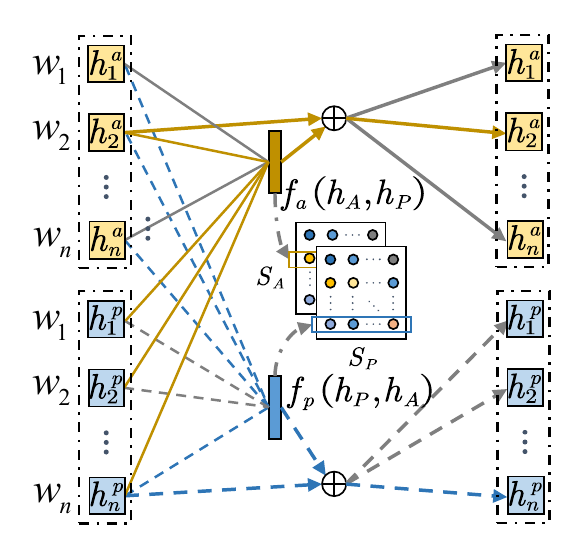}
			\label{fig_cross_shared_unit}
		}
		\caption{An illustration of the proposed DOER framework.}\label{fig_framework}
	\end{figure*}

	The proposed framework is shown in Figure \ref{fig_framework_dcs}. We will first formulate the aspect term-polarity co-extraction problem and then describe this framework in detail in this section.

	\subsection{Problem Statement}
	\label{sec_problem}
	This paper deals with aspect term-polarity co-extraction, in which the aspect terms are explicitly mentioned in the text. We solve it as two sequence labeling tasks. Formally, given a review sentence $S$ with $n$ words from a particular domain, denoted by $S=\{w_i | i = 1, \dots, n\}$. For each word $w_i$, the objective of ATE is to assign it a tag $t^a_i \in T^a$, and likewise, the objective of ASC is to assign a tag $t^p_i \in T^p$, where $T^a=\{$B, I, O$\}$ and $T^p=\{$PO, NT, NG, CF, O$\}$. The tags B, I and O in $T^a$ stand for the beginning of an aspect term, the inside of an aspect term, and other words, respectively. The tags PO, NT, NG, and CF indicate polarity categories: positive, neutral, negative, and conflict, respectively. The tag O in $T^p$ means other words like that in $T^a$. Figure \ref{table_labeling_examples} shows a labeling example of the first sentence in Figure \ref{fig_examples}.
	
	\subsection{Model Overview}
	\label{sec_model_overview}
	
	We discuss the proposed framework DOER in detail below.
	
	\vspace{+1.5mm}
	\noindent
	\textbf{Word Embedding.} \quad Instead of adopting standard techniques to generate the embedding of each word $w_i$ by concatenating word embedding and char embedding, we use the double embeddings proposed in \cite{Xu2018} as the initial word embeddings. The double embeddings contain two types: general-purpose embeddings and domain-specific embeddings, which are distinguished by whether the embeddings are trained by an in-domain corpus or not. Formally, each word $w_i$ will be initialized with a feature vector $h_{w_i} \in \mathbb{R}^{d_G+d_D}$, where $d_G$ and $d_D$ are the first dimension size of the general-purpose embeddings $\mathrm{G} \in \mathbb{R}^{d_G \times |V|}$ and the domain-specific embeddings $\mathrm{D} \in \mathbb{R}^{d_D \times |V|}$, respectively. $|V|$ is the size of the vocabulary. Hence, $h_{w_i}$ is generated by $h_{w_i} = G(w_i) \oplus D(w_i)$, where $\oplus$ means the concatenation operation. $h_g$ and $h_d$ in Figure \ref{fig_framework_dcs} denote $G(w_i)$ and $D(w_i)$, respectively. All the out-of-vocabulary words are randomly initialized, and all sentences are padded (or tailored when testing) and initialized with zeros to the max length of the training sentences.
	\begin{figure}[htbp]
		\centering
		\includegraphics[width=0.34\textwidth, keepaspectratio]{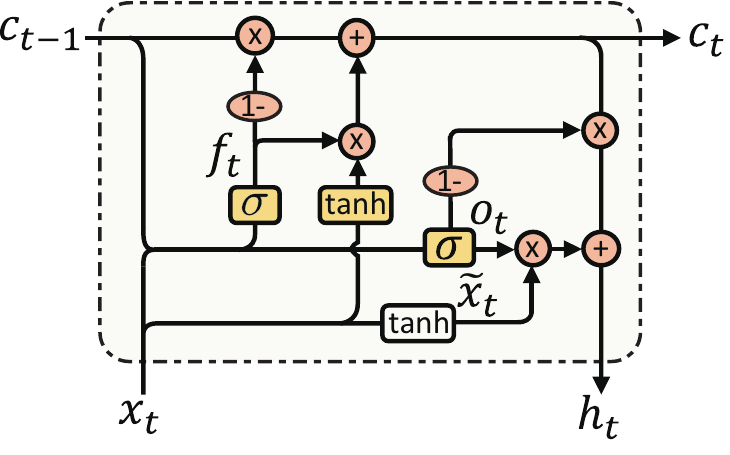}
		\caption{Residual gated unit (ReGU).}\label{fig_regu}
	\end{figure}

	\vspace{+1.5mm}
	\noindent
	\textbf{Stacked Dual RNNs.} \quad The main architecture of DOER is a stacked dual RNNs, one stacked RNN for ATE, and one stacked RNN for ASC. Each layer of RNNs is a bidirectional ReGU (BiReGU). As shown in Figure \ref{fig_regu}, ReGU has two gates to control the flow of input and hidden state. Given input $x_t$ at time $t$ and the previous memory cell $c_{t-1}$, the new memory cell $c_t$ is calculated via the following equation:
	\begin{align}
	c_t = (1-f_t) \odot c_{t-1} + f_t \odot \tanh(W_ix_t),
	\end{align}
	and the new hidden state $h_t$ is then computed as
	\begin{align}
	h_t = (1-o_t) \odot c_t + o_t \odot \tilde{x}_t, \label{eq_mgru_h}
	\end{align}
	where $f_t = \sigma\left( W_fx_t + U_fc_{t-1} \right)$ is a forget gate, $o_t = \sigma\left( W_ox_t + U_oc_{t-1} \right)$ is a residual gate, and $\tilde{x}_t$ is $x_t$ or $\tanh(W_xx_t)$ according to whether the size of $x_t$ is equal to $c_t$ or not. $f_t$ controls the information flow from the previous timestamp to the next timestamp. $o_t$ controls the information flow from the previous layer to the next layer. $\sigma$ denotes the logistic function, $tanh$ means the hyperbolic tangent function, and $\odot$ is element-wise multiplication. $W_*$ of size $d \times d_{I}$ and $U_*$ of size $d \times d$ are weight matrices, where $*\in\{i, f, o, x\}$. The bias vectors are omitted for simplicity. The size of $d_I$ changes with the dimension of the input. Its value is $d_G+d_D$ when it is the first layer of the stacked BiReGU.
	
	BiReGU owns two directional representations of the input like Bidirectional LSTM \cite{Graves2005b}. We concatenate the hidden states generated by ReGU in both directions belonging to the same input as the output vector, which is expressed as $h_{t} = \overrightarrow{h}_t \oplus \overleftarrow{h}_t$, where $\oplus$ again means concatenation. $\overrightarrow{h}_t$ and $\overleftarrow{h}_t$ have the same formulation as Eq. (\ref{eq_mgru_h}) but different propagation directions. Thus, the size of $h_{t}$ is $2d$, and the size of $d_I$ will also become $2d$ when stacking a new BiReGU layer. We refer the outputs of dual BiReGU as $h_A$ and $h_P$ separately to differentiate ATE and ASC.
	
	\vspace{+1.5mm}
	\noindent
	\textbf{Cross-Shared Unit.} \quad When generating the representation after BiReGU, the information of ATE and ASC is separated from each other. However, the fact is that the labels of ATE and the labels of ASC have strong relations. For instance, if the label of ATE is O, the label for ASC should be O  as well, and if the label of ASC is PO, the label for ATE should be B or I. Besides, both the labels of ATE and the labels of ASC have the information to imply the boundary of each aspect term.
	
	The cross-shared unit (CSU) is used to consider the interaction of ATE and ASC. We first compute the composition vector $\alpha_{ij}^M \in \mathbb{R}^K$ through the following tensor operator:
	\begin{align}
		\alpha_{ij}^M = f_m\left( h_i^m, h_j^{\overline{m}} \right) = \tanh\left( (h_i^m)^{\top}G^mh_j^{\overline{m}} \right),
	\end{align}
	where $M \in \{A, P\}$, $m \in \{a, p\}$, $h_i^m \in h_M$, and $G^m \in \mathbb{R}^{K \times 2d \times 2d}$ are 3-dimensional tensors. $K$ is a hyperparameter. $A, a$ and $P, p$ are indexes of ATE and ASC, respectively, $\overline{m}=p, M=A$ if $m=a$, and $\overline{m}=a, M=P$ if $m=p$. Such tensor operators can be seen as multiple bilinear terms, which have the capability of modeling more complicated compositions between two vectors \cite{Socher2013,Wang2017}. 

	After obtaining the composition vectors, the attention score $S_{ij}^M$ is calculated as:
	\begin{align}
		S_{ij}^M = v_m^{\top}\alpha_{ij}^M,
	\end{align}
	where $v_m \in \mathbb{R}^K$ is a weight vector used to weight each value of the composition vector, $M \in \{A, P\}$, and $m \in \{a, p\}$. Thus, $S_{ij}^M$ is a scalar. All these scalars $S_{ij}^A$ and $S_{ij}^P$ are gathered in two matrices $S_A$ and $S_P$, respectively. A higher score $S_{ij}^A$ indicates a higher correlation between aspect term $i$ and the polarity representation captured from $j$-th word. Likewise, a higher score $S_{ij}^P$ indicates a higher correlation between aspect polarity $i$ and the representation of aspect term captured from $j$-th word. We use their related representations to enhance the original ATE and ASC features through:
	\begin{align}
		h_M = h_M + \text{softmax}_r\left( S^M \right)h_{\overline{M}}, 
	\end{align}
	where $\text{softmax}_r$ is a row-based softmax function, $M \in \{A, P\}$, $\overline{M}=P$ if $M=A$, and $\overline{M}=A$ if $M=P$. Such an operation can make ATE and ASC get enhanced information from each other. The process is shown in Figure \ref{fig_cross_shared_unit}.

\vspace{+1.5mm}
	\noindent
	\textbf{Interface.} \quad To generate the final ATE tags and ASC tags, either a dense layer plus a softmax function or a Conditional Random Fields (CRF) can be used. According to the comparison in \cite{Reimers2017a}, using a CRF instead of a softmax classifier as the last layer can obtain a performance increase for tasks with a high dependency between tags. Thus, we use the linear-chain CRF as our inference layer. Its log-likelihood is computed as follows:
	\begin{align}
	\label{eq_crf}
	L\left( W_c,b_c \right) =\sum_i{\log\ p \left( y|h;W_c,b_c \right)}.
	\end{align}
	where $p \left( y|h;W_c,b_c \right)$ is the probability function of CRF, and $W_c$ and $b_c$ are the weight and bias, respectively. The Viterbi algorithm is used to generate the final labels of ATE and ASC.
	
	\vspace{+1.5mm}
	\noindent
	\textbf{Joint Output.} \quad After generating the labels for ATE and ASC in the inference layer, the last step is to obtain the aspect term-polarity pairs. It is convenient to get the aspect terms of the given sentence according to the meaning of the elements in $T^a$. To generate the polarity of each aspect term, we use the aspect term as the boundary of polarity labels, and then count the number of each polarity category within the boundary and adopt the label that has the maximum number or the first label (if all the numbers of each polarity category are equal) as the final polarity. For example, the final polarity of ``PO NT'' is ``PO'', the final polarity of ``PO PO'' is also ``PO'', and the final polarity of ``PO NT NT'' is ``NT''. This method is simple and effective in our experiments.
	
	\vspace{+1.5mm}
	\noindent
	\textbf{Auxiliary Aspect Term Length Enhancement.} \quad Although CRF is capable of considering the correlation of two adjacent labels, there are generated discontinuous labels, especially for a long target aspect term. To alleviate the influence resulted from the length of the aspect term, we designed an auxiliary task to predict the average length of aspect terms in each sentence when training the model. The computational process of the prediction in ATE is as follows:
	\begin{align}
		z_{u_A} = \sigma \left( W_{u_A}^{\top}\tilde{h}_A \right),
	\end{align}
	where $\tilde{h}_A \in \mathbb{R}^{2d}$ is the result of max-pooling of $h_A^{l_1}$, which is generated by the first RNN layer, $W_{u_A} \in \mathbb{R}^{2d}$ is a weight parameter. We calculate the prediction loss through the mean squared error (MSE):
	\begin{align}
		\mathcal{L}_{u_A} = \lVert z_{u_A} - \hat{z}_{u} \rVert^2,
	\end{align}
	where $\hat{z}_{u}$ is the average length of aspect terms in a sentence after global normalization on the training dataset.

	ASC has a similar prediction process to ATE after the first layer of the stacked RNNs, but it has different weight $W_{u_P}$ and hidden feature $\tilde{h}_P$ than $W_{u_A}$ and $\tilde{h}_A$. The prediction loss is denoted by $\mathcal{L}_{u_P}$.

    \vspace{+1.5mm}
	\noindent
	\textbf{Auxiliary Sentiment Lexicon Enhancement.} \quad As previously discussed, the polarity of an aspect term is usually inferred from its related opinion words. Thus, we also use a sentiment lexicon to guide ASC. Specifically, we train an auxiliary word-level classifier on the branch of ASC for discriminating positive words and negative words based on the sentiment labels $\hat{Y}_p^S$. This means that we use a sentiment lexicon to map each word of a sentence to a sentiment label in training. For each feature of ASC $h_i^{p,l_1}$ generated by the first RNN layer, we use a linear layer and the softmax function to get its sentiment label:
	\begin{align}
		z_i^s = \text{softmax} \left( W_s^{\top}h_i^{p,l_1} \right),
	\end{align}
	where $W_s \in \mathbb{R}^{2d \times c}$ is a weight parameter, $c=3$ means the sentiment label is one of the three elements in the set $\{$positive, negative, none$\}$. We use the cross-entropy error to calculate the loss of each sentence:
	\begin{align}
		\mathcal{L}_s = -\frac{1}{n}\sum_{i=1}^n \left( \mathbb{I} \left(\hat{y}_i^S \right) \left( \text{log} \left(z_i^s \right) \right)^\top \right),
	\end{align}
	where $\mathbb{I}(\hat{y}_i^S)$ means the one-hot vector of $\hat{y}_i^S \in \hat{Y}_p^S$.

	\subsection{Joint Loss}
	\label{sec_training}
	On the whole, the proposed framework DOER has two branches: one for ATE labeling and the other for ASC labeling. Each of them is differentiable, and thus can be trained with gradient descent. We equivalently use the negative of $L\left( W_c,b_c \right)$ in Eq. (\ref{eq_crf}) as the error to do minimization via back-propagation through time (BPTT) \cite{Goller1996}. Thus, the loss is as follows:
	\begin{align}
	\label{eq_loss}
	\mathcal{L} = -\sum_i{\log\ p \left( y|h;W_c,b_c \right)},
	\end{align}
	Then, the losses from both tasks and the auxiliary tasks are constructed as the joint loss of the entire model:
	\begin{align}
		\mathcal{J}(\Theta) \!\!=\!\! (\mathcal{L}_a \!\!+\!\! \mathcal{L}_p) \!\!+\!\! (\mathcal{L}_{u_A} \!\!+\!\! \mathcal{L}_{u_P} \!\!+\!\! \mathcal{L}_{s}) \!\!+\!\! \frac{\lambda}{2} \lVert \Theta \rVert^2,
	\end{align}
	where $\mathcal{L}_a$ and $\mathcal{L}_p$, which have the same formulation as Eq. (\ref{eq_loss}), denote the loss for aspect term and polarity, respectively. $\Theta$ represents the model parameters containing all weight matrices $W$, $U$, $v$ and bias vectors $b$. $\lambda$ is a regularization parameter.
	\begin{table}[tp]
		\begin{center}
			\begin{tabular}{|cl|l|l|ll|}
				\hline
				\multicolumn{2}{|c|}{Datasets}    & Train & Dev & \multicolumn{1}{l|}{Test} & Total \\ \hline
				\multirow{4}{*}{$\mathbb{S}_\text{L}$}      & \#PO &  941     & 32    & \multicolumn{1}{l|}{340}     & 1,313      \\
											 & \#NT &  446     &  4   & \multicolumn{1}{l|}{169}     & 619      \\
											 & \#NG &  820     &  17   & \multicolumn{1}{l|}{126}     & 963      \\
											 & \#CF &  41     &  1   & \multicolumn{1}{l|}{16}     & 58      \\ \hline
				\multirow{4}{*}{$\mathbb{S}_\text{R}$} & \#PO & 3,262      & 126    & \multicolumn{1}{l|}{1,490}     & 4,878      \\
											 & \#NT &  674     &  13   & \multicolumn{1}{l|}{250}     & 937      \\
											 & \#NG &  1,205     & 46    & \multicolumn{1}{l|}{500}     & 1,751      \\ 
											 & \#CF &  88     &  0   & \multicolumn{1}{l|}{14}     & 102      \\ \hline
				\multirow{3}{*}{$\mathbb{S}_\text{T}$}     & \#PO & \multicolumn{3}{c|}{-}                   & 698      \\
											 & \#NT & \multicolumn{3}{c|}{-}                   & 2,254      \\
											 & \#NG & \multicolumn{3}{c|}{-}                   & 271      \\ \hline
			\end{tabular}
		\end{center}
		\caption{\label{table-datasets} Datasets from SemEval and Twitter.}
	\end{table}
	
	\section{Experiments}
	\subsection{Datasets}
	We conduct experiments on two datasets from the SemEval challenges and one English Twitter dataset. The details of these benchmark datasets are summarized in Table \ref{table-datasets}. $\mathbb{S}_\text{L}$ comes from SemEval 2014 \cite{Pontiki2014}, which contains laptop reviews, and $\mathbb{S}_\text{R}$ are restaurant reviews merged from SemEval 2014, SemEval 2015 \cite{Pontiki2015}, and SemEval 2016 \cite{Pontiki2016}. We keep the official data division of these datasets for the training set, validation set, and testing set. The reported results of $\mathbb{S}_\text{L}$ and $\mathbb{S}_\text{R}$ are averaged scores of 10 runs. $\mathbb{S}_\text{T}$ consists of English tweets. Due to lack of standard train-test split, we report the ten-fold cross-validation results of $\mathbb{S}_\text{T}$ as done in \cite{Mitchell2013,Zhang2015,Li2019}. For the auxiliary task of sentiment lexicon enhancement, we exploit a sentiment lexicon~\footnote{\url{http://mpqa.cs.pitt.edu/} (the lexicon of~\cite{Hu2004}~\url{https://www.cs.uic.edu/~liub/FBS/sentiment-analysis.html} can be used as well.} to generate the label when training the model. The evaluation metric is F1 score based on the exact match of aspect term and its polarity.
	
	\subsection{Word Embeddings}
	To initialize the domain-specific word embeddings, we train the word embeddings by CBOW \cite{Mikolov2013} using Amazon reviews\footnote{\url{http://jmcauley.ucsd.edu/data/amazon/}} and Yelp reviews\footnote{\url{https://www.yelp.com/academic_dataset}}, which are in-domain corpora for laptop and restaurant respectively. Thus, for $\mathbb{S}_\text{L}$, we use Amazon embedding, and for $\mathbb{S}_\text{R}$, we use Yelp embedding. The Amazon review dataset contains 142.8M reviews, and the Yelp review dataset contains 2.2M restaurant reviews. The embeddings from all these datasets are trained by Gensim\footnote{\url{https://radimrehurek.com/gensim/}} which contains the implementation of CBOW. The parameter \emph{min\_count} is set to 10 and \emph{iter} is set to 200. We use Amazon embedding as the domain-specific word embeddings of $\mathbb{S}_\text{T}$ as Amazon corpora is large and comprehensive although not in the same domain. The general-purpose embeddings are initialized by Glove.840B.300d embeddings \cite{Pennington2014}. Its corpus is crawled from the Web.

	\subsection{Settings}
	In our experiments, the regularization parameter $\lambda$ is empirically set as $0.001$, and $d_G$ and $d_D$ as 300 and 100, respectively. The hidden state size of $d$ of ReGU is 300. The hyperparameter $K$ is set to 5. We use Adam \cite{Kingma2015} as the optimizer with the learning rate of 0.001 and the batch size of 16. We also employ dropout \cite{Srivastava2014} on the outputs of the embedding layer and two BiReGU layers. The dropout rate is 0.5. To avoid the exploding gradient problem, we clip the gradient norm within 5. The maximum number of epochs is set to 50. The word embeddings are fixed during the training process. We implemented DOER using the TensorFlow library \cite{Abadi2016a}, and all computations are done on an NVIDIA Tesla K40 GPU.

	\subsection{Baseline Methods}
	To validate the performance of the proposed model DOER~\footnote{The code of DOER is available at \url{https://github.com/ArrowLuo/DOER}} on the aspect term-polarity co-extraction task, a comparative experiment is conducted with the following baseline models:
	\begin{itemize}
		\item \textbf{CRF-\{pipelined, joint, collapsed\}}: They leverage linguistically informed features with CRF to perform the sequence labeling task using the pipelined, joint, or collapsed approach\footnote{\url{http://www.m-mitchell.com/code/}} \cite{Mitchell2013}.
		\item \textbf{NN+CRF-\{pipelined, joint, collapsed\}}: An improvement of \cite{Mitchell2013} that concatenates target word embedding and context four-word embeddings besides using linguistically informed features plus CRF to finish the sequence labeling task \cite{Zhang2015}. Instead of using the officially released code\footnote{\url{https://github.com/SUTDNLP/OpenTargetedSentiment}} due to the outdated library, we reproduce the results with the original settings.
		\item \textbf{Sentiment-Scope}: A collapsed CRF model\footnote{\url{https://github.com/leodotnet/sentimentscope}} \cite{Li2017}, which expands the node types of CRF to capture sentiment scopes. The discrete features used in this model are exactly the same as the above two groups of models.
		\item \textbf{DE-CNN+TNet}: DE-CNN\footnote{\url{https://github.com/howardhsu/DE-CNN}} \cite{Xu2018} and TNet \cite{Li2018b} are the current state-of-the-art models for ATE and ASC, respectively. DE-CNN+TNet combines them in a pipelined manner. We use the official TNet-AS variant\footnote{\url{https://github.com/lixin4ever/TNet}} as our TNet implementation.
		\item \textbf{LSTM+CRF-\{LSTMc, CNNc\}}: They all use BiLSTM plus CRF for sequence labeling. The difference is that LSTM+CRF-LSTMc \cite{Lample2016} encodes char embedding by BiLSTM, while LSTM+CRF-CNNc \cite{Ma2016} uses CNN.
		\item \textbf{LM-LSTM-CRF}: It is a language model enhanced LSTM-CRF model proposed in \cite{Liu2018}, which achieved competitive results on several sequence labeling tasks\footnote{\url{https://github.com/LiyuanLucasLiu/LM-LSTM-CRF}}.
		\item \textbf{E2E-TBSA}: It is an end-to-end model of the collapsed approach proposed to address ATE and ASC simultaneously\footnote{\url{https://github.com/lixin4ever/E2E-TBSA}} \cite{Li2019}.
		\item \textbf{S-BiLSTM}: It is a stacked BiLSTM model with two layers that adopts the joint approach and has the same Embeddings, Interface, Joint Output layers as DOER.
		\item \textbf{S-BiReGU}: It is similar to S-BiLSTM but uses a ReGU cell instead of an LSTM cell.
	\end{itemize}

	\begin{table*}[tp]
		\begin{center}
			\begin{tabular}{|ll|p{1.7cm}<{\centering}|p{1.7cm}<{\centering}|p{1.7cm}<{\centering}|}
				\hline
				 & \textbf{Model} & $\mathbb{S}_\text{L}$ & $\mathbb{S}_\text{R}$ & $\mathbb{S}_\text{T}$ \\ \hline 
				\multirow{3}{*}{\textbf{Pipeline Baselines}}  & CRF-pipeline           & 51.08 & 54.78 & 31.91 \\
				& NN+CRF-pipeline        & 53.36 & 60.78 & 45.08 \\
				& DE-CNN+TNet            & 56.47 & 67.54 & 48.74 \\ \hline
				\multirow{7}{*}{\textbf{Collapsed Baselines}} & CRF-collapsed         & 49.24 & 59.52 & 32.00 \\
				& NN+CRF-collapsed       & 50.64 & 61.74 & 45.52 \\
				& Sentiment-Scope        & 50.27  & 62.01  & 45.91  \\
				& LSTM+CRF-LSTMc         & 54.43 & 65.93 & 46.57 \\
				& LSTM+CRF-CNNc          & 54.71 & 66.36 & 47.35 \\
				& LM-LSTM-CRF            & 56.39 & 67.56 & 48.46 \\
				& E2E-TBSA               & 57.99 & 69.91 & 49.13 \\ \hline
				\multirow{2}{*}{\textbf{Joint Baselines}}     & CRF-joint              & 50.73 & 59.75 & 32.42 \\
				& NN+CRF-joint           & 52.81 & 60.27 & 44.69 \\ \hline
				\multirow{6}{*}{\textbf{Ours}}                & S-BiLSTM         & 56.83 & 71.22 & 48.94 \\ 
				& S-\textbf{BiReGU}         & 57.82 & 71.47 & 49.11 \\
				& S-\textbf{BiReGU}+\textbf{CSU}     & 58.99 & 72.19 & 49.89 \\
				& S-\textbf{BiReGU}+\textbf{CSU}+\textbf{AuL} & 59.06 & 72.32 & 51.06 \\
				& S-\textbf{BiReGU}+\textbf{CSU}+\textbf{AuS} & 60.11 & 72.64 & 51.13 \\
				& \textbf{DOER}                   & \textbf{60.35} & \textbf{72.78} & \textbf{51.37} \\ \hline
				\end{tabular}
		\end{center}
		\caption{\label{table-results-compare} F1 score (\%) comparison of all systems for aspect term-polarity pair extraction.}
	\end{table*}
	We use two abbreviations AuL and AuS for the ablation study. \textbf{AuL} denotes the auxiliary task of aspect term length enhancement, and \textbf{AuS} denotes the auxiliary task of sentiment lexicon enhancement. All baselines have publicly available codes, and we ran these officially released codes to reproduce the baseline results except the NN+CRF variants due to the outdated library as discussed in the bullet point for these baseline systems.

	\subsection{Results and Analysis}
	\noindent
	\textbf{Comparison Results.} \quad The comparison results are shown in Table \ref{table-results-compare}, which are F1 scores of aspect term-polarity pairs. As the results show, our DOER obtains consistent improvement over baselines. Compared to the best pipelined model, the proposed framework outperforms DE-CNN+TNet by 3.88\%, 5.24\%, and 2.63\% on $\mathbb{S}_\text{L}$, $\mathbb{S}_\text{R}$, and $\mathbb{S}_\text{T}$, respectively. It indicates that an elaborated joint model can achieve better performance than pipeline approaches on aspect term-polarity co-extraction task. Besides, seven collapsed models are also introduced to the comparison. Compared to the best of these collapsed approaches, DOER improves by 2.36\%, 2.87\%, and 2.24\% over E2E-TBSA on $\mathbb{S}_\text{L}$, $\mathbb{S}_\text{R}$, and $\mathbb{S}_\text{T}$, respectively. This result shows the potential of a joint model which considers the interaction between the two relevant tasks. Comparing with existing works based on the joint approach, i.e., CRF-joint and NN+CRF-joint, DOER makes substantial gains over them as well. The improvements over DE-CNN+TNet and E2E-TBSA are statistically significant ($p<0.05$).
	
	\vspace{+1.5mm}
	\noindent
	\textbf{Ablation Study.} \quad To test the effectiveness of each component of DOER, we conduct an ablation experiment with results shown in the last block of Table \ref{table-results-compare}. The fact that S-BiReGU gives superior performance compared to S-BiLSTM indicates the effectiveness of ReGU in our task. This residual architecture enables information transfer to the next layers more effective. With the help of CSU, S-BiReGU+CSU achieves better performance than without it. We believe the interaction of information between ATE and ASC is essential to improve each other. Although the samples with long aspect terms are rare, the auxiliary task of aspect term length can improve the performance. Another auxiliary task of sentiment lexicon can also enhance the representation of the proposed framework. As a whole of S-BiReGU, CSU, AuL, and AuS, the proposed DOER achieves superior performance. It mainly benefits from the enhanced features by the two auxiliary tasks and the interaction of two separate routes of ATE and ASC.

	\vspace{+1.5mm}
	\noindent
	\textbf{Results on ATE.} \quad Table \ref{table-results-ate} shows the results of aspect term extraction only. DE-CNN is the current state-of-the-art model on ATE as mentioned above. Comparing with it, DOER achieves new state-of-the-art scores. DOER$^*$ denotes the DOER without ASC part. As the table shows, DOER achieves better performance than DOER$^*$, which indicates the interaction between ATE and ASC can yield better performance for ATE than only conduct a single task.
	\begin{table}[htbp]
		\begin{center}
			\begin{tabular}{|l|p{1.4cm}<{\centering}|p{1.4cm}<{\centering}|p{1.4cm}<{\centering}|}
				\hline
				\textbf{Model}     & $\mathbb{S}_\text{L}$ & $\mathbb{S}_\text{R}$ & $\mathbb{S}_\text{T}$ \\ \hline 
				DE-CNN             & 81.26 & 78.98 & 63.23  \\ 
				DOER$^*$           & 82.11 & 79.98 & 68.99  \\ 
				DOER               & 82.61 & 81.06 & 71.35 \\ \hline
				\end{tabular}
		\end{center}
		\caption{\label{table-results-ate} F1 score (\%) comparison only for aspect term extraction.}
	\end{table}

	\vspace{+1.5mm}
	\noindent
	\textbf{Case Study.} \quad Table \ref{table-case-study} shows some examples of S-BiLSTM, S-BiReGU+CSU, and DOER. As observed in the first and second rows, S-BiReGU+CSU and DOER predict the aspect term-polarity pair correctly but S-BiLSTM does not. With the constraint of CSU, the error words can be avoided as shown in the second row. The two auxiliary tasks work well on the CSU. They can capture a better sentiment representation, e.g., the third row, and alleviate the misjudgment on the long aspect terms, e.g., the last row.

\begin{figure}[tp]
		\centering
		\includegraphics[width=0.31\textwidth, keepaspectratio]{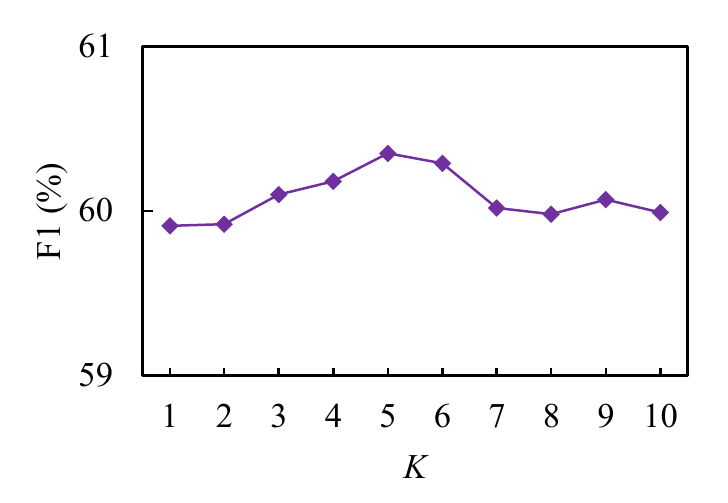}
		\caption{F1 scores on $\mathbb{S}_\text{L}$ with different $K$.}
		\label{fig_parameter_k}
	\end{figure}

	\vspace{+1mm}
	\noindent
	\textbf{Impact of $K$.} \quad We investigate the impact of hyperparameter $K$ of the CSU on the final performance. The experiment is conducted on $\mathbb{S}_\text{L}$ by varying $K$ from 1 to 10 with the step of 1. As shown in Figure \ref{fig_parameter_k}, value 5 is the best choice for the proposed method to address our task. Due to the performance demonstrated in the figure, $K$ is set to 5 cross all experiments for simplicity.
	\begin{table*}[tp]
		\begin{center}
			\begin{tabular}{m{4.8cm}|m{3.3cm}|m{3.2cm}|m{3.2cm}}
				\hline
				\textbf{Input} & \multicolumn{1}{c|}{\textbf{S-BiLSTM}} & \multicolumn{1}{c|}{\textbf{S-BiReGU+CSU}} & \multicolumn{1}{c}{\textbf{DOER}} \\ \hline \hline
				I like the [\textcolor{mypink}{lighted screen}]$_\text{PO}$ at night. & None (\xmark) & [lighted screen]$_\text{PO}$ & [lighted screen]$_\text{PO}$ \\ \hline 
				It is a great [\textcolor{mypink}{size}]$_\text{PO}$ and amazing [\textcolor{mypink}{windows 8}]$_\text{PO}$ included! & $\text{[size]}_\text{PO}$, [windows 8 included]$_\text{PO}$ (\xmark) &  $\text{[size]}_\text{PO}$, $\text{[windows 8]}_\text{PO}$ & $\text{[size]}_\text{PO}$, $\text{[windows 8]}_\text{PO}$ \\ \hline 
				I tried several [\textcolor{mypink}{monitors}]$_\text{NT}$ and several [\textcolor{mypink}{HDMI cables}]$_\text{NT}$ and this was the case each time. & [HDMI cables]$_\text{NG}$ (\xmark) & None (\xmark), [HDMI cables]$_\text{NT}$ & [monitors]$_\text{NT}$, [HDMI cables]$_\text{NT}$ \\ \hline
				The [\textcolor{mypink}{2.9 ghz dual-core i7 chip}]$_\text{PO}$ really out does itself. & [dual-core i7 chip]$_\text{PO}$ (\xmark) & [dual-core i7 chip]$_\text{PO}$ (\xmark) & [2.9 ghz dual-core i7 chip]$_\text{PO}$ \\ \hline
			\end{tabular}
		\end{center}
		\caption{\label{table-case-study} Case analysis on S-BiLSTM, S-BiReGU+CSU, and DOER. \ding{55} means wrong prediction.}
	\end{table*}

\begin{figure}[htbp]
		\centering
		\subfloat[$S_A$] {
			\centering
			\includegraphics[width=0.20\textwidth, keepaspectratio]{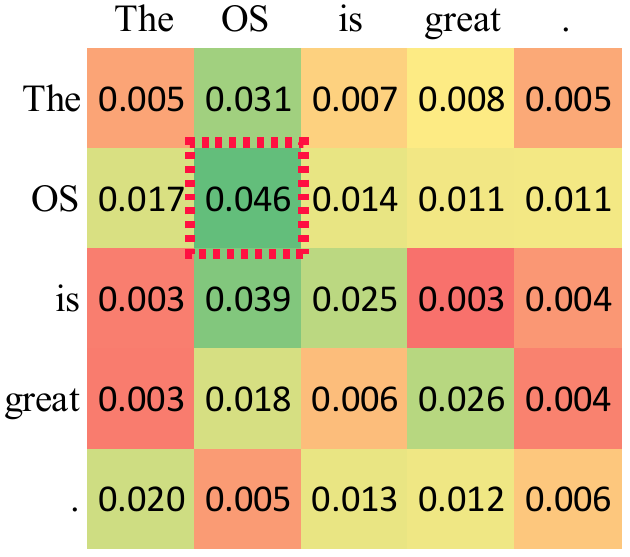}
			\label{fig_CSU_attentionA}
		}
		\subfloat[$S_P$] {
			\centering
			\includegraphics[width=0.20\textwidth, keepaspectratio]{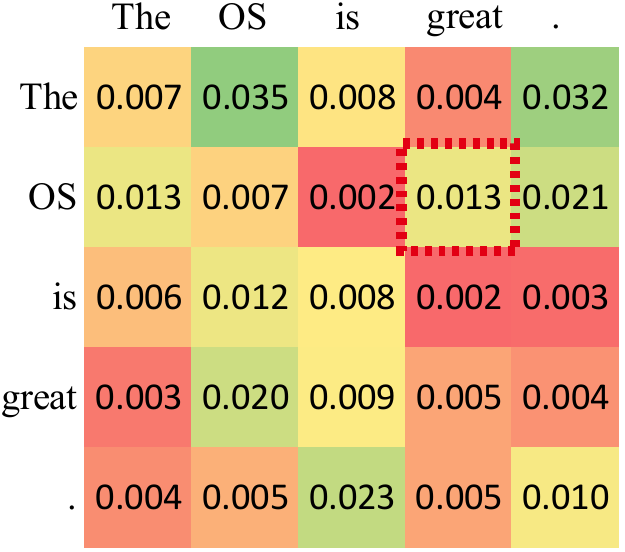}
			\label{fig_CSU_attentionP}
		}
		\caption{Visualization of $S_A$ and $S_P$ in CSU.}
		\label{fig_CSU_attention}
	\end{figure}

	\vspace{+1mm}
	\noindent
	\textbf{Visualization of Attention Scores in CSU.} \quad We also try to visualize the attention scores $S_A$ and $S_P$ to explore the effectiveness of CSU. As shown in Figure \ref{fig_CSU_attention}, $S_A$ and $S_P$ have different values, which indicate that both ATE and ASC indeed interact with each other. The red dashed rectangle in Figure \ref{fig_CSU_attentionA} shows that the model learns to focus on itself when labeling the word ``OS'' in the ATE task. Likewise, the red dashed rectangle in Figure \ref{fig_CSU_attentionP} shows that the model learns to focus on the word ``great'' instead of itself when labeling the word ``OS'' in the ASC task. The fact that the polarity on the target aspect ``OS'' is positive, which is inferred from the ``great'', verifies that the system is doing the right job. In summary, we can conclude that the attention scores learned by CSU benefit the labeling process.

	\section{Related Work}
	\label{sec_related_work}
	Our work spans two major topics of aspect-based sentiment analysis: aspect term extraction and aspect sentiment classification. Each of them has been studied by many researchers. \citeauthor{Hu2004} (\citeyear{Hu2004}) extracted aspect terms using frequent pattern mining. \citeauthor{Qiu2011} (\citeyear{Qiu2011}) and \citeauthor{Liu2015Automated} (\citeyear{Liu2015Automated}) proposed to use rule-based approach exploiting either hand-crafted or automatically generated rules about some syntactic relationships. \citeauthor{Mei2007} (\citeyear{Mei2007}), \citeauthor{He2011auto} (\citeyear{He2011auto}) and \citeauthor{Chen2014Aspect} (\citeyear{Chen2014Aspect}) used topic modeling based on Latent Dirichlet Allocation \cite{Blei2003}. All of the above methods are unsupervised. For supervised methods, the ATE task is usually treated as a sequence labeling problem solved by CRF. For the ASC task, a large body of literature has tried to utilize the relation or position between the aspect terms and the surrounding context words as the relevant information or context for prediction \cite{Tang2016a,Arjun2016extract}. Convolution neural networks (CNNs) \cite{Poria2016aspect,Li2018}, attention network \cite{Wang2016attention,Ma2017a,He2017}, and memory network \cite{Wang2018a} are also active approaches.
	
	However, the above methods are proposed for either the ATE or the ASC task.	\citeauthor{Lakkaraju2014Aspect} (\citeyear{Lakkaraju2014Aspect}) proposed to use hierarchical deep learning to solve these two subtasks. \citeauthor{Wu2016} (\citeyear{Wu2016}) utilized cascaded CNN and multi-task CNN to address aspect extraction and sentiment classification. Their main idea is to directly map each review sentence into pre-defined aspect terms by using classification and then classifying the corresponding polarities. We believe the pre-defined aspect terms are in general insufficient for most analysis applications because they will almost certainly miss many important aspects in review texts.
	
	This paper regards ATE and ASC as two parallel sequence labeling tasks and solves them simultaneously. Comparing with the methods that address them one by one using two separate models, our framework is easy to use in practical applications by outputting all the aspect term-polarity pairs of input sentences at once. Similar to our work, \citeauthor{Mitchell2013} (\citeyear{Mitchell2013}) and \citeauthor{Zhang2015} (\citeyear{Zhang2015}) are also about performing two sequence labeling tasks, but they extract named entities and their sentiment classes jointly. We have a different objective and utilize a different model. \citeauthor{Li2019} (\citeyear{Li2019}) have the same objective as us. The main difference is that their approach belongs to a collapsed approach but ours is a joint approach. The model proposed by \cite{Li2017} is also a collapsed approach based on CRF. Its performance is heavily dependent on manually crafted features.
	
	\section{Conclusion}
	In this paper, we introduced a co-extraction task involving aspect term extraction and aspect sentiment classification for aspect-based sentiment analysis and proposed a novel framework DOER to solve the problem. The framework uses a joint sequence labeling approach and focuses on the interaction between two separate routes for aspect term extraction and aspect sentiment classification. To enhance the representation of sentiment and alleviate the difficulty of long aspect terms, two auxiliary tasks were also introduced in our framework. Experimental results on three benchmark datasets verified the effectiveness of DOER and showed that it significantly outperforms the baselines on aspect term-polarity co-extraction. 

	\section*{Acknowledgments}
	This work is supported by the National Key R\&D Program of China (No. 2017YFB1401401).

	
	\bibliographystyle{acl_natbib}
	\bibliography{Co-AspectExtraction-PolarityClassification-Ref}
\end{document}